\documentclass[final,12pt]{clear2023} 

\title{Scalable Causal Discovery with Score Matching}

\usepackage{mathtools}      
\DeclarePairedDelimiter{\abs}{\lvert}{\rvert} 
\usepackage{times}
\usepackage{graphicx} 
\usepackage{centernot}
\usepackage{tikz}
\usepackage{caption}
\usepackage{subcaption}
\usepackage{booktabs}
\usepackage{multirow}
\usepackage{setspace} 
\usepackage{enumitem} 
\usepackage{wrapfig}

\newcommand{\indep}{\,\rotatebox[origin=c]{90}{$\models$}\,} 

\DeclarePairedDelimiterX{\expectarg}[1]{[}{]}{%
  \ifnum\currentgrouptype=16 \else\begingroup\fi
  \activatebar#1
  \ifnum\currentgrouptype=16 \else\endgroup\fi
}
\newcommand{\innermid}{\nonscript\;\delimsize\vert\nonscript\;}
\newcommand{\activatebar}{%
  \begingroup\lccode`\~=`\|
  \lowercase{\endgroup\let~}\innermid 
  \mathcode`|=\string"8000
}

\clearauthor{%
 \Name{Francesco Montagna}\thanks{Work has been partially carried out during an internship at Amazon Web Services, Tubingen} \Email{francesco.montagna@edu.unige.it}\\
    \addr MaLGa-DIBRIS, Università di Genova
 \AND
  \Name{Nicoletta Noceti} \Email{nicoletta.noceti@unige.it}\\
  \addr MaLGa-DIBRIS, Università di Genova
 \AND
  \Name{Lorenzo Rosasco} \Email{lrosasco@mit.edu}\\
  \addr MaLGa-DIBRIS, Università di Genova\\
  MIT, CBMM\\
  Istituto Italiano di Tecnologia
 \AND
  \Name{Kun Zhang} \Email{kunz1@cmu.edu}\\
  \addr Carnegie Mellon University\\
  MBZUAI
 \AND
 \Name{Francesco Locatello} \Email{locatelf@amazon.com}\\
 \addr AWS
}

\begin{document}
\maketitle

\begin{abstract}
\looseness=-1This paper demonstrates how to discover the whole causal graph from the second derivative of the log-likelihood in observational nonlinear additive Gaussian noise models. Leveraging scalable machine learning approaches to approximate the score function $\nabla \operatorname{log}p(\mathbf{X})$, we extend the work of~\citet{rolland_2022} that only recovers the topological order from the score and requires an expensive pruning step removing spurious edges among those admitted by the ordering.
Our analysis leads to DAS (acronym for Discovery At Scale), a practical algorithm that reduces the complexity of the pruning by a factor proportional to the graph size. In practice, DAS achieves competitive accuracy with current state-of-the-art while being over an order of magnitude faster. Overall, our approach enables principled and scalable causal discovery, significantly lowering the compute bar.
\end{abstract}


\section{Introduction}
\label{sec:intro}
Causal discovery from observational data is a central problem affecting virtually all scientific domains, such as biology, genetics, economics, and machine learning (\cite{sachs_2005, koeller_daphne_friedman, pearl_2009, PetersJanzingSchoelkopf17}). Given a causal model one can predict the effect of interventions on the system's variables without the need of accessing interventional data which might be costly, unfeasible or unethical to collect. On the other hand, inferring causal relations from observational data is generally non-identifiable and requires additional assumptions.  
In traditional causality research, algorithms to  discover causal relationships from observations can be divided in three classes (\cite{kun_clark_peter,scholkopf2021toward}). \textit{Constraint-based} approaches like PC (\cite{pc_alg}), FCI and SGS (\cite{Spirtes2000}) test the conditional independence between the variables and search for graphs structures that satisfies them under a faithfulness assumption. Usually they do not output a unique graph but an equivalence class. The main bottleneck of these approaches is that conditional independence testing is notoriously difficult (\cite{Shah_2020}). \textit{Score-based} methods define a suitable score function, and search for the graph that best fits the data. Greedy approaches, such as GES (\cite{chickering_2003}), are used to search in this large space, which size grows super-exponentially with the number of nodes (thus limiting scalability). Finally, a \textit{restricted model} class assumption, e.g., nonlinear relations and additive Gaussian noise, allows to identify the Directed Acyclic Graph (DAG) underlying the observations (\cite{PetersJanzingSchoelkopf17,B_hlmann_2014,grandag_19, notears_2018, shimizu06}).\\[.5em]
One main challenge affecting the discovery of the causal graph is that enforcing the DAG constraint has a cubic per-iteration cost in the number of variables, making the optimization the computational bottleneck. One approach to reduce the computational requirements, is to decouple the causal discovery task in two steps: first, a topological ordering is found, such that a node can be a parent only of its successors in the ordering, thus enforcing the acyclicity constraint. Then, a pruning step selects the correct subset of edges among those admitted by the inferred ordering, removing all spurious connections in the graph.
In this setting, a step towards better scalability is the work of \citet{rolland_2022} that recently proposed the SCORE algorithm: first they efficiently estimate the score function $\nabla \operatorname{log}p(\mathbf{X})$, then they recover the topological order from the Jacobian of the score, and finally they prune the fully connected DAG by the method proposed in CAM (\cite{B_hlmann_2014}). The pruning step is the bottleneck of SCORE, amounting to $95\%$ of the runtime on graphs with $50$ nodes and scaling cubic in the number of nodes.
\begin{wrapfigure}{l}{0.48\textwidth}
\begin{center}
    \includegraphics[width=.45\textwidth]{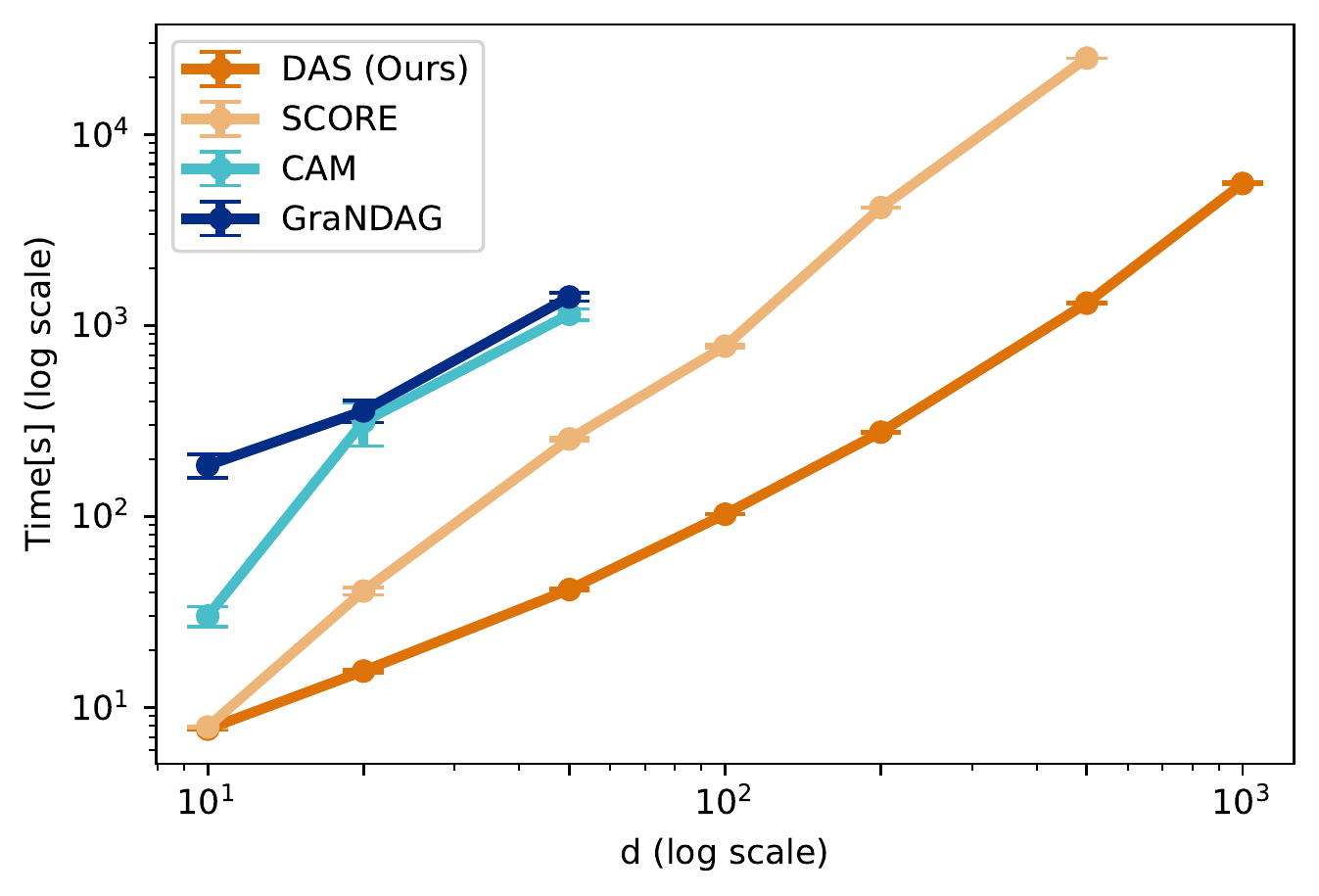}
    \caption{Execution time of different methods versus $d$ number of nodes for dense graphs (ER4 data).}
    \label{fig:time}
\end{center}
\end{wrapfigure}

\looseness-1In this work, we show that the second derivative of the log-likelihood allows to recover both the skeleton and the direction of the edges in the causal DAG. Theoretically, this implies that we can get rid of the costly pruning step in SCORE~(\cite{rolland_2022}) as all information about the causal structure is already contained in the Jacobian of the score. While our analysis yields a practical algorithm, we found it beneficial to first identify few candidate edges and still retain a final cheap pruning step. This is now much more efficient as most of the edges have already been detected and it is only needed to correct mistakes from the finite samples approximation of the score, reducing the complexity by a factor proportional to the number of nodes in the graph. This advantage is clearly visible in practice, reducing the runtime of SCORE by over an order of magnitude as shown in Figure~\ref{fig:time}. \\
Our contributions can be summarized as follows:
\begin{itemize}
    \item We demonstrate how to theoretically recover the full causal DAG from the score of the data distribution. This extends prior work showing that the topological order can be recovered from the score (\cite{rolland_2022}).
    \item We introduce DAS\footnote{The code for DAS is available as part of the DoDiscover library \url{https://www.pywhy.org/dodiscover/dev/index.html}} (acronym for Discovery At Scale), an algorithm for efficient and scalable causal discovery. As we lower the computational bar to apply causal discovery techniques on large numbers of variables, we also present clear examples and highlight when our algorithm is expected to fail. We expect these to be useful for practitioners interested in using DAS on their data as well as researchers working on scalable causal discovery.
    \item Our analysis yields a practical approach to filter the candidate edges in the final pruning step. While our method is marginally less accurate than~(\cite{rolland_2022}) it improves the runtime by an order of magnitude in the graph size. We demonstrate the speedup improvement on synthetic graphs with up to a thousand nodes. 
\end{itemize}

\section{Background knowledge}

We introduce the background needed for our analysis starting from the formalism of structural causal models.

\subsection{Structural Causal Models}
\label{sec:scm}
One way to formalize causal relationships between variables is with an additive Structural Causal Model (SCM). Consider a set $\mathbf{X}=\{X_i\}_{i=1}^{d}$ of observable  vertices of a DAG. We assume that the structure of the graph can be expressed in the functional relationship
\begin{equation}
    X_i = f_i(\operatorname{pa}_i(\mathbf{X})) + \epsilon_i, \hspace{5mm} \epsilon_i \sim \mathcal{N}(0, \sigma_i), \hspace{5mm} \forall i = 1,  \ldots, d \hspace{2mm},
    \label{eq:model}
\end{equation}
with $\operatorname{pa}_i(\mathbf{X})$ set of parent nodes of $X_i$ in the directed network. We will assume $X_i \in \mathbb{R}$, additive and independently drawn Gaussian noise elements $\epsilon_i$, as well as $f_i$ to be twice continuously differentiable and nonlinear in every component.\\[.5em]
\looseness-1 Recursive application of \eqref{eq:model} allows to derive the joint probability distribution $p(X_1, X_2, \ldots, X_d)$. As this probability is over vertices of a directed acyclic graph, the following factorization holds (\cite{pearl_2009, lauritzen1996}):
\begin{equation}
    p(\mathbf{X}) = \prod_{i=1}^d p(X_i|\operatorname{pa}_i(\mathbf{X})) \hspace{2mm}.
    \label{eq:p_factorization}
\end{equation}
The additive noise model \eqref{eq:model} is known to be identifiable under \textit{mild} assumptions (\cite{hoyer2009ANM, Zhang09_additive}), meaning that we can recover uniquely the causal graph from observational data generated according to the joint distribution over $\mathbf{X}$.

\paragraph{Problem definition} \looseness-1Usually the form of the $f_i$ in the model \eqref{eq:model} is not known and neither is the probability in \eqref{eq:p_factorization}, while we can only access a set of observations from the joint distribution. Given these observations the task is to identify the causal structure of the graph underlying the SCM. This problem is known as causal discovery. One solution is to use data to estimate a topological ordering of the variables in $\mathbf{X}$, and then to choose edges of the DAG between those admitted by such ordering. In our approach we select edges that satisfy constraints derived from the Jacobian of the score.

\subsection{Topological ordering of a graph} \label{sec:top_order}
Given a set of nodes $\mathbf{X}=\{X_i\}_{i=1}^{d}$, let $\mathcal{G} = (\mathbf{X}, \mathcal{E})$ be a DAG. A topological order relative to $\mathcal{G}$ is a permutation $\mathbf{X}^{\pi}$ of the nodes such that an edge $(i,j)$ in $\mathcal{G}$ implies $X_i$ appearing before $X_j$ in $\mathbf{X}^{\pi}$, denoted as $X_j \succ X_i$. Intuitively, a node can be a parent only of its successors in the ordering. According to this definition more than one topological ordering is allowed for a given DAG. On the other hand, there is a one to one correspondence between a given permutation $\mathbf{X}^{\pi}$ and a fully connected graph $\mathcal{G^{\pi}}$ where we draw edges $(i, j)$ for each $i, j = 1, \ldots , d$ such that $X_j \succ X_i$ in $\mathbf{X^\pi}$. If $\mathbf{X}^\pi$ is equal to $\mathbf{X}^*$, one of the correct permutations of the real DAG $\mathcal{G}$, then $\mathcal{G}^\pi = (\mathbf{X}, \mathcal{E}^\pi)$ is a supergraph of $\mathcal{G}$ meaning that its set of edges satisfies $\mathcal{E} \subseteq \mathcal{E}^\pi$. In the next section we provide an overview of CAM edges selection procedure, that allows to correctly identify $\mathcal{E}$ given a topological ordering $\mathbf{X}^{\pi}$ and its corresponding dense set $\mathcal{E}^\pi$.

\vspace{.5em}

\subsection{Preliminary Neighbours Search (PNS) and CAM-pruning}\label{sec:cam}
Now we briefly describe the two-steps pruning method of CAM (\cite{B_hlmann_2014}), namely \textit{Preliminary Neighbours Search} (PNS) followed by \textit{CAM-pruning}, which is used to remove \textit{spurious} edges of $\mathcal{G}^\pi$, the dense graph induced by a permutation $\mathbf{X}^\pi$.

\paragraph{PNS}
PNS is a neighbours selection method for nonlinear additive Gaussian noise models, following the idea of variable selection in graphs presented in \cite{Meinshausen_2006} for the linear Gaussian case. In particular, given an order $\mathbf{X}^\pi$, variable selection is performed by fitting for each $j = 1, \ldots, d$ an additive model of $X_j$ versus all the other variables $\{X_i: X_j \succ X_i \textnormal{ in } \mathbf{X}^\pi\}$, and choosing the $K$ most \textit{important} predictor variables as possible parents of $X_j$. 
This restricts the number of possible incoming edges of $X_j$ to an arbitrary fixed amount $K$, reducing the total complexity of the edges search procedure and the workload of CAM-pruning step. PNS is implemented by the authors with a boosting method for additive models fitting (\cite{B_hlmann_2007, B_hlmann_2003}). The total complexity of PNS is therefore $\mathcal{O}(dt\operatorname{r}(n,d))$ as for each node $1, \ldots, d$ the boosting algorithm fits $t$ models of complexity $\operatorname{r}(n,d)$, with $n$ the number of samples in the dataset. The complexity term $\operatorname{r}(n,d)$ depends on the choice of the additive model fitting technique, amounting to $\mathcal{O}(nd^2)$ using Iteratively Reweighted Least Squares (\cite{minka2003comparison}).

\paragraph{CAM-pruning}
After \textit{Preliminary Neighbours Search}, a final pruning step is performed by applying significance testing of covariates to remove \textit{superfluous} edges and thus reducing the number of false positives. In this case the computational complexity is negligible being bounded by the fixed parameter $K$ decided in PNS.\\
\looseness=-1Our work aims at replacing PNS with a novel approach making CAM-pruning application scalable to high dimensional graphs. In particular we reduce asymptotic complexity of the edge search procedure from $\mathcal{O}(nd^3)$ to $\mathcal{O}(d^2)$.  

\section{Deducing causal structure from the score}
For the causal discovery problem under analysis we consider an observable $\mathbf{X} \in \mathbb{R}^d$ whose entries $X_i$ are vertices of a graph generated according to the model in \eqref{eq:model}. In the next section, we show how the score function is in principle sufficient to solve this task. First we illustrate the ideas behind the SCORE algorithm, which can estimate the score's Jacobian and find a topological ordering of the variables of interest. Then we derive additional constraints on the Jacobian matrix of the score that allows to identify the edges of the causal graph. 

\subsection{SCORE overview}
\looseness-1\cite{rolland_2022} introduces a method for efficiently estimating the score function $s(\mathbf{X})$ and its Jacobian exploiting the Stein identity (\cite{Stein1972ABF}). Complementary to this, they propose a method to identify leaf nodes in a causal graph generated according to \eqref{eq:model} by inspection of the diagonal elements of the Jacobian of the score. \\
In order to derive the leaf identification procedure, first we need to find a closed form of $s(\mathbf{X}) = \nabla \log p(\mathbf{X})$. Starting from Equation \eqref{eq:p_factorization}, we have:
\begin{equation}
\begin{split}
    \operatorname{log}p(\mathbf{X}) &= \sum_{i=1}^{d}\operatorname{log}p(X_i | \operatorname{pa}_i(\mathbf{X})) = \\
    &= -\frac{1}{2} \sum_{i=1}^{d} \left(\frac{X_i - f_i(\operatorname{pa}_i(\mathbf{X}))}{\sigma_i^2} \right)^2 -\frac{1}{2} \sum_{i=1}^{d} \operatorname{log}(2\pi \sigma_i)^2 \hspace{2mm}.
\end{split}
\end{equation}
The $j$-th entry of $\nabla \operatorname{log}p(\mathbf{X})$ therefore is
\begin{equation}
    s_j(\mathbf{X}) = -\frac{X_j - f_j(\operatorname{pa}_j(\mathbf{X}))}{\sigma_j^2} + \sum_{i \in \operatorname{ch}_j{(\mathbf{X}})} \frac{\partial f_i}{\partial x_j}(\operatorname{pa}_i(\mathbf{X}))\hspace{1mm} \frac{X_i - f_i(\operatorname{pa}_i(\mathbf{X}))}{\sigma_i^2} \hspace{2mm},
    \label{eq:score_entry}
\end{equation}
\looseness-1with $ch_j(\mathbf{X})$ denoting the set of children of node $X_j$. Now, let $X_j$ be a leaf node: being the set of children nodes $ch_j(\mathbf{X}) = \emptyset$, from Equation \eqref{eq:score_entry} we notice that the score simplifies as follow:
\begin{equation}
    s_j(\mathbf{X}) = -\frac{X_j - f_j(\operatorname{pa}_j(\mathbf{X}))}{\sigma_j^2}.
    \label{eq:leaf_score}
\end{equation}
\looseness-1Moreover, it is easy to verify that $\frac{\partial s_j(\mathbf{X})}{\partial x_j} = -\frac{1}{\sigma^2_j}$, such that the diagonal entry of the score's Jacobian associated to a leaf node is a constant.
Based on this relation, Lemma 1 of \cite{rolland_2022} provides a formal criterion to identify leaves in a causal graph. Given a node $X_i$, the following holds:
\begin{equation}
    \frac{\partial s_i(\mathbf{X})}{\partial x_i} = c \Longleftrightarrow X_i \textnormal{ is a leaf, } \: \forall i = 1, \ldots, d,
    \label{eq:lemma_rolland}
\end{equation}
where $c \in \mathbb{R}$ is a constant scalar value. This relation directly implies that
\begin{equation}
    \operatorname{Var}\left[\frac{\partial s_i(\mathbf{X})}{\partial x_i} \right] = 0 \Longleftrightarrow X_i \textnormal{ is a leaf, } \: \forall i = 1, \ldots, d.
    \label{eq:var_lemma_rolland}
\end{equation}
In order to find the complete topological ordering, SCORE algorithm of \cite{rolland_2022} is designed as follow: first it estimates the Jacobian of the score $\hat{J}(s(\mathbf{X}))$, that is used to identify a leaf in the graph by \eqref{eq:var_lemma_rolland}. Then, it removes the leaf from the graph, assigning it a position in the order vector. By iteratively repeating this two steps procedure up to the source node, all variables in $\mathbf{X}$ end up being assigned a position in the causal ordering.\\[.5em]
In the following section we show that given that a topological ordering is known, we can derive additional constraints on the off-diagonal elements of the score's Jacobian that identify directed edges in the graph.

\subsection{Deriving constraints for edge selection}
\looseness-1We observe that for a leaf node $l$, $X_l \in \mathbf{X}$, the partial derivative of \eqref{eq:score_entry} over $X_j$ with $j \neq l$ is:
\begin{equation}
\frac{\partial s_l(\mathbf{X})}{\partial X_j} = 
    \begin{cases}
    \frac{1}{\sigma_l^2} \frac{\partial f_l}{\partial X_j}(\operatorname{pa}_l(\mathbf{X})) \neq 0 & \textnormal{if $X_j \in \operatorname{pa}_l(\mathbf{X})$ } \\[1em]
    0 & \textnormal{else}
    \end{cases}  \hspace{3mm}.
\label{eq:method}
\end{equation}
\looseness-1It is worth to notice that $\frac{1}{\sigma_l^2} \frac{\partial f_l}{\partial X_j}(\operatorname{pa}_l(\mathbf{X}))$ might still be vanishing for some values of $\operatorname{pa}_l(\mathbf{X})$ even if $X_j \in \operatorname{pa}_l(\mathbf{X})$, for instance if the function has a maximum or a minimum: given the assumption on $f_l$ nonlinear even when considered on a restricted interval, these events happen with probability zero, such that $\frac{1}{\sigma_l^2} \frac{\partial f_l}{\partial X_j}(\operatorname{pa}_l(\mathbf{X})) \neq 0$ holds \textit{almost surely}. We prove that the condition in Equation \eqref{eq:method} allows to derive a criterion to identify parents of a given leaf node by slightly adapting the result of~\cite{rolland_2022}.
\begin{lemma}[Adapted from~\cite{rolland_2022}]
\label{lem:lemma1}
Let $p$ be the probability density function of a random variable $\mathbf{X} \in \mathbb{R}^d$ defined via nonlinear additive Gaussian noise model \eqref{eq:model}. Let also $s(\mathbf{X}) = \nabla \operatorname{log}p(\mathbf{X})$ be the associated score function. Without loss of generality, assume a topological ordering  $\mathbf{X}^{\pi} = (X_1, \ldots, X_d)$. Then given a leaf $l$:
\begin{equation}
    \mathbf{E}\left[ \abs*{\frac{\partial s_l(\mathbf{X})}{\partial X_j}}\right] \neq 0 \Longleftrightarrow X_j \in \operatorname{pa}_l(\mathbf{X}), \:\: \forall j \in \{1, \ldots, l-1\} \:.
\end{equation}
\end{lemma}
The proof is provided for completeness in the Appendix \ref{app:proof}.

\paragraph{Difference of Lemma~\ref{lem:lemma1} with \cite{rolland_2022}} The formulation in \cite{rolland_2022} requires $\operatorname{Var}\left[\frac{1}{\sigma_l^2} \frac{\partial f_l}{\partial X_j}(\operatorname{pa}_l(\mathbf{X}))\right] \neq 0 \Leftrightarrow X_j \in \operatorname{pa}_l(\mathbf{X})$, where $X_l$ is a leaf node. We illustrate the problem with this considering a simple two variables case with graph $X_1 \longrightarrow X_2$: if parent node $X_1$ has zero variance, their selection condition would break, predicting a graph with $X_1$ and $X_2$ independent. While this case would be ruled out by the assumption of variance larger than zero for every node, in practice it can be a problem. Given a finite sample $X \in \mathbb{R}^{n \times d}$ and its topological ordering $\mathbf{X}^{\pi}$, if parents of a leaf $X_l$ show small variance in the sample, we might still mistake the oscillation observed in $\frac{1}{\sigma_l^2} \frac{\partial f_l}{\partial X_j}(\operatorname{pa}_l(\mathbf{X}))$ for statistical error due to finite set estimates, discarding an existing edge.
\looseness-1To estimate the edges we need higher statistical accuracy compared to the topological order since in the former case we do not know how many parents a node has while in the latter we know there is always at least one leaf. This is why we adapted the Lemma of~\cite{rolland_2022} to a theoretically equivalent but practically more robust formulation. 
We rely on the sample mean of the absolute value of the score's Jacobian entries $\frac{1}{\sigma_l^2} \frac{\partial f_l}{\partial X_j}(\operatorname{pa}_l(\mathbf{X}))$ for the implementation of Lemma \ref{lem:lemma1}: this estimator is potentially subject to the same issues, but shows better robustness properties than the sample variance (due to the absolute value) and estimating a lower moment yields lower error (estimating variance requires estimating the mean first, so any statistical error in the mean estimator affects the variance estimator), making it a preferable choice. 

\vspace{.7em}
In practice we can exploit Lemma \ref{lem:lemma1} to reconstruct the entire graph only if an ordering $\mathbf{X}^{\pi}$ is provided. To see why, consider the last entry $X_l$ of $\mathbf{X^\pi}$: by definition of topological ordering $X_l$ is a leaf. Then we can apply Lemma \ref{lem:lemma1} doing partial derivatives of $s_l(\mathbf{X})$ over all nodes $\{X_j : X_j \prec X_l \textnormal{ in } \mathbf{X}^\pi\}$ and identify as parents those that satisfy the required constraint. At this point, we remove $X_l$ from the ordering $\mathbf{X}^\pi$ and repeat the procedure on the pruned graph with vertices $\mathbf{X} \setminus X_l$. By iterating these steps over each node in the ordering from last to source we can identify the exact graph. 

\begin{example}
To clarify these ideas we discuss a simple three variables example illustrating how the results in Lemma \ref{lem:lemma1} enables edges discovery. \\[.7em]
Let $\mathbf{X} = (X_1, X_2, X_3)$ with $X_i \in \mathbb{R}$ generated according to model \eqref{eq:model}. Consider the topological ordering $\mathbf{X}^\pi = (X_3, X_2, X_1)$ to be given. The goal is to recover the real causal graph with $X_3$ as source node, $pa_1(\mathbf{X}) = \{X_2\}$ and $pa_2(\mathbf{X}) = \{X_3\}$.

\begin{figure}[h!]
    \centering
    \label{fig:example}
    \begin{subfigure}{0.4\textwidth}
    \subcaption{Graph $\mathcal{G}^\pi$ admitted by $\mathbf{X}^\pi$}
        \begin{tikzpicture}
            \node[shape=circle,draw=black] (A) at (0,0) {$X_3$};
            \node[shape=circle,draw=black] (B) at (-1,-1) {$X_2$};
            \node[shape=circle,draw=black] (C) at (0,-2) {$X_1$};
            
            \draw[->] (A) -- (B);
            \draw[->] (A) -- (C);
            \draw[->] (B) -- (C);
        \end{tikzpicture}
    
    \end{subfigure}
    \begin{subfigure}{0.4\textwidth}
        \subcaption{Correct causal graph of $\mathbf{X}$}
        \begin{tikzpicture}
            \node[shape=circle,draw=black] (A) at (0,0) {$X_3$};
            \node[shape=circle,draw=black] (B) at (-1,-1) {$X_2$};
            \node[shape=circle,draw=black] (C) at (0,-2) {$X_1$};
            
            \draw[->] (A) -- (B);
            \draw[->] (B) -- (C);
        \end{tikzpicture}
    
    \end{subfigure}
    \label{fig:my_label}
\end{figure}

We proceed analyzing the Jacobian of the score function for $X_1$ and $X_2$, while we easily see from the ordering $\mathbf{X}^\pi$ that $X_3$ has no parents.

\begin{itemize}
    \item $X_1$: according to \eqref{eq:leaf_score} its score is simply $s_1(\mathbf{X}) = -\frac{X_1 - f_1(X_2)}{\sigma_1^2}$. Its off diagonal partial derivatives are
    \begin{equation*}
        \begin{split}
            \frac{\partial s_1}{\partial X_2}(\mathbf{X}) &= \frac{1}{\sigma_1^2}\frac{\partial f_1}{\partial X_2}(X_2) \hspace{2mm}, \\
            \frac{\partial s_1}{\partial X_3}(\mathbf{X}) &= 0 \hspace{2mm}.
        \end{split}
    \end{equation*}
    Applying Lemma \ref{lem:lemma1} we correctly conclude that the set of parents of $X_1$ is $\operatorname{pa}_1(\mathbf{X}) = \{X_2\}$.
    
    \item $X_2$: in order to apply Lemma \ref{lem:lemma1} also to $X_2$ we need to discard $X_1$ from the graph, such that $X_2$ is a leaf node. Given the set of variables $\mathbf{\Tilde{X}} = \mathbf{X} \setminus \{X_1\}$ we get the graph of Figure \ref{fig:g_step2}. Then $s_2(\mathbf{\Tilde{X}}) = -\frac{X_2 - f_2(X_3)}{\sigma_2^2}$ and the partial derivative is
    \begin{equation*}
        \frac{\partial s_2}{\partial X_3}(\mathbf{\Tilde{X}}) =  \frac{1}{\sigma_2^2}\frac{\partial f_2(X_3)}{\partial X_3} \hspace{2mm}.
    \end{equation*}
    By Lemma \ref{lem:lemma1} we see that $\operatorname{\mathbf{E}}\left[ \abs*{\frac{1}{\sigma_2^2}\frac{\partial f_2(X_3)}{\partial X_3}}\right] \neq 0$, such that $X_3$ is a parent of $X_2$.
    
    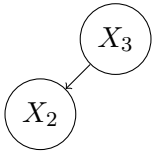
\begin{figure}
        \centering
        \caption{Causal graph of $\mathbf{\Tilde{X}} = (X_2, X_3)$.}
        \label{fig:g_step2}
        \begin{tikzpicture}
            \node[shape=circle,draw=black] (A) at (0,0) {$X_3$};
            \node[shape=circle,draw=black] (B) at (-1,-1) {$X_2$};
            
            \draw[->] (A) -- (B);
        \end{tikzpicture}
    \end{figure}
\end{itemize}

The algorithm recovers the exact structure.

\end{example}

\paragraph{Discussion}
These findings on identifiability of the causal structure from the score function are not completely surprising in the light of previous results on Markov networks (\cite{spantini_2017, morrison_17}). 
Given a collection of random variables $\mathbf{X} = (X_1, X_2, \ldots, X_d)$ with joint density $p(\mathbf{X})$, the information of conditional independencies between the variables of $\mathbf{X}$ can be embedded in a simple undirected Markov network $\mathcal{G} = (\mathcal{V}, \mathcal{E})$, where edges $(i, j)$ encode some sort of probabilistic interaction between the pairs of random variables $X_i,X_j$. 
In particular \citet{spantini_2017} proved how to construct a Markov graph reading the conditional independence of pairs of random variables as follow:
\begin{equation}
\label{eq:cond_indep}
    X_j \indep X_i \,|\, \mathbf{X}_{ \mathcal{V} \setminus\{i,j\}} \Longleftrightarrow \partial_{ij} \log p(\mathbf{X}) = 0 \hspace{2mm},
\end{equation}
where $\partial_{ij}(\cdot)$ denotes the $ij$-th mixed partial derivative and $\partial_{ij} \log p(\mathbf{X})$ is an entry of the Jacobian of the score. By adding edges between each couple of nodes that appears not to satisfy Equation \eqref{eq:cond_indep}, we obtain an undirected graph encoding all and only the existing conditional independencies between the variables of $\mathbf{X}$. \\[.5em]
Equation \eqref{eq:method} of our work discovers the same constraint in a slightly different setting: rather than evaluating $\partial_{ij} \log p(\mathbf{X})$ for each node against every other, we follow an iterative approach where first we identify a leaf $X_l$ and then we test its mixed derivatives only against nodes coming before in the topological ordering. By the time we find an edge we know its direction as we know that $X_l$ is a leaf, which breaks the symmetry in the relation. Moreover Lemma \ref{lem:lemma1} ensures correct identification of directed \textit{v-structures} like $i \rightarrow j \leftarrow k$ that instead in the conditional independence map are \textit{moralized} with an additional link $(i, k)$, thus allowing to recover all edges and their direction.

\vspace{.7em}
Next we derive an algorithm for causal discovery based on Lemma \ref{lem:lemma1}, and show how it retains performance with respect to other state of the art methods, while showing better scalability in the number of nodes.

\subsection{DAS: an algorithm for causal Discovery At Scale}
\label{sec:das}
We want to use the constraint of Lemma \ref{lem:lemma1} on the score function to derive an algorithm for causal discovery which is faster and exhibits better scaling properties in the number of nodes than any other technique to our knowledge. Given a set $X \in \mathbb{R}^{n \times d}$ of $n$ observations generated according to model \eqref{eq:model}, first we estimate a topological ordering $\hat{\mathbf{X}}^\pi$ via the SCORE algorithm. Then, we iterate over each node $X_l$ of such ordering starting from the last, which we know to be a leaf, and identify incoming edges of $X_l$ as follow: to begin we estimate the score's Jacobian $\hat{J}$ via the SCORE algorithm; we have now $n$ estimates of the Jacobian of the score, one for each of the $n$ data points. According to Lemma \ref{lem:lemma1}, we consider the absolute value of the $l$-th row of the $n$ Jacobian matrices, and look for entries with non-zero mean: this can be achieved by statistical hypothesis testing, where the idea is to test for the mean of a sample to be different from zero. In practice, we perform hypothesis testing according to the following heuristic method: we compute the average of the absolute value of the entries in the $l$-th row, and select as potential parents of $X_l$ the $K+1$ nodes associated to largest averages. This parameter $K$ is the same that we find in $PNS$ algorithm, which limits the maximum number of nodes fed to the pruning algorithm. Between these $K+1$ entries, we find a reference node $X_{ref}$ whose samples of $\abs*{\hat{J}_{l, ref}}$ have the average closest to zero: at this point, for each of the remaining $K$ nodes, we perform Welchs's t-test with the null hypothesis of equality of the population means $H_0: \: \operatorname{\mathbf{E}}[\abs*{J_{l,j}}] = \operatorname{\mathbf{E}}[\abs*{J_{l,ref}}]$ and the alternative $H_1: \: \operatorname{\mathbf{E}}[\abs*{J_{l,j}}] > \operatorname{\mathbf{E}}[\abs*{J_{l,ref}}]$, with $X_j$ potential parent of $X_l$. If we reject the null with \textit{p-value} $\leq 0.01$, then $X_j$ is added to the parents of $X_l$ in an adjacency matrix representing the inferred graph.\\
Eventually, $X_l$ column is removed from the matrix of the data $X$, and the procedure is repeated for another leaf node found in the ordering. Once every node in $\mathbf{X}^\pi$ is considered, we prune the resulting adjacency matrix via \textit{CAM-pruning} (Section \ref{sec:cam}), simply with the goal of reducing the number of false positives. \\
The implementation details of DAS are illustrated in the pseudo-code of Algorithm \ref{alg:das}.

\subsection{Algorithmic complexity}\label{sec:complexity}
Considering an input matrix $n \times d$ with $n$ the number of samples and $d$ the number of nodes, the overall complexity of DAS is $\mathcal{O}(dn^3 + d^2)$. Indeed estimating the topological order with SCORE involves inverting a $n \times n$ matrix for $d$ times, one for each iteration necessary to identify a leaf node: hence, the $\mathcal{O}(dn^3)$ contribute. Additionally the edge search step requires iterating over the $d$ elements of the ordering, each time selecting the $K$ largest entries on a list of size $\leq d$ (see Algorithm \ref{alg:das}) yielding a $\mathcal{O}(d^2)$ contribution.\\
\looseness-1On the other hand SCORE, arguably the most scalable state-of-the-art algorithm for causal discovery, uses PNS and CAM-pruning to select edges while shares the same ordering of DAS. The bottleneck in SCORE 
execution is the preliminary neighbours search step, whose complexity has been studied in detail in Section \ref{sec:cam} and amounts to $\mathcal{O}(dt \operatorname{r}(n, d))$ with $\operatorname{r}(n,d)$ the number of operations to fit a generalized additive model ( $\mathcal{O}(n d^2)$~(\cite{minka2003comparison}) with Iteratively Reweighted Least Squares). Therefore our use of the score function for candidate edges selection dramatically improves the execution time allowing to scale causal discovery in high dimensions by a factor of $\mathcal{O}(d)$.
\begin{algorithm}
\footnotesize
\caption{DAS}\label{alg:das}
\setstretch{1.4}
Input: data matrix $X \in \mathbb{R}^{n \times d}$, $K \in \mathbb{R}$

$X^{\pi} \leftarrow \operatorname{SCORE}(X)$ \hspace{6em}($X^{\pi}[d]$ leaf node)

$X^{\pi} \leftarrow reverse(X^\pi)$ \hspace{5.63em}($X^{\pi}[d]$ source node)

$A \leftarrow d \times d$ zeros adjacency matrix

\For{$X_l$ in $X^{\pi}$}{
$\hat{J} \leftarrow \left[\abs*{\frac{\partial s_l}{\partial X_j}}\right]_{X_j \prec X_l \textnormal{ in } X^\pi} $ \hspace{1.5em} (estimate from SCORE)
    
    $K \leftarrow \operatorname{min}(K, \operatorname{length}(\hat{J}))$
    
    $topK = topK(\operatorname{Average}(\hat{J}))$ \hspace{1.4em}($\,$topK($\cdot$) : return indices of the K largest values of the input)
    
    $j_{ref} = topK[\operatorname{argmin} \hat{J}[topK]]$
    
    \For{\textit{k} \textnormal{in} \textit{topK}}{
    $p = $ p-value for the test $H_0: \: \mu_{J_k} = \mu_{J_{ref}}$, $H_1: \: \mu_{J_k} > \mu_{J_{ref}}$ \hspace{2em}($\mu$ = mean)
    
        \If{$p < 0.01$}{
        $A[k, l] = 1$
        }
    }
    
    Remove $l$-th column from $X$
}    
$\hat{\mathcal{G}} \leftarrow \textnormal{CAM-pruning}(A)$

\Return $\hat{\mathcal{G}}$
\end{algorithm}

\paragraph{Non identifiability of the linear model}
Next we further highlight the consistency of our algorithm by showing how it fails in identifying the causal graph under the assumption of linear $f_j$ in model \eqref{eq:model}, $\forall j = 1, \ldots, d$. Indeed it has been proven that observational data generated according to a linear additive Gaussian noise model do not allow for recovery of the underlying causal structure (\cite{PetersJanzingSchoelkopf17, COMON1994287}) unless additional assumptions are made (\cite{Peters_2013}). In the simplest setting of two variables $X, Y$ linked in a causal graph, unidentifiability amounts to the impossibility of choosing the edge direction, i.e., it is not possible to decide whether $X$ is the cause or the effect of $Y$.\\ 
Given that the topological ordering between variables itself disambiguates the direction of edges in the DAG, we have to show that our topological ordering method on a set $\mathbf{X} \in \mathbb{R}^d$ fails in the linear setting.  
SCORE algorithm identifies leaf nodes by finding terms with zero variance in the diagonal of the score's Jacobian, as specified in Equation \eqref{eq:var_lemma_rolland}. In case of an SCM as defined in \eqref{eq:model} but with linear functions $f_j$, we find that $\frac{\partial s_j}{\partial X_j}(\mathbf{X}) = -\frac{1}{\sigma_j^2} + \operatorname{constant}$ for every node $j$ in the graph. Thus it can be easily seen that
\begin{equation}
\label{eq:score_fail}
    \operatorname{Var}\left[\frac{\partial s_j}{\partial X_j}(\mathbf{X})\right] = 0, \: \forall j = 1, \ldots, d \hspace{2mm}.
\end{equation} 
Since the variance in \eqref{eq:score_fail} vanishes for each node $j$ rather than for leaves only, then the criterion of Equation \eqref{eq:var_lemma_rolland} does not hold anymore. This implies failure of the topological ordering method for the linear case in accordance with our claim.\\[.5em]
In the next section we study the algorithmic complexity of DAS and we highlight the better efficiency with respect to SCORE.

\section{Experiments}\label{sec:exp}
Now we summarize experimental outcomes of DAS method in comparison with several state of the art algorithms for causal discovery. We report results of SCORE-ordering with PNS and CAM-pruning steps (named simply SCORE in the table) (\cite{rolland_2022}), CAM \footnote{CAM refers to both topological ordering and pruning steps introduced in the original paper. PNS is applied only for $d \geq 50$.} (\cite{B_hlmann_2014}) and GraNDAG (\cite{grandag_19}). Other algorithms such as PC and FCI are omitted as they perform much worse (\cite{B_hlmann_2014, grandag_19}). Up to $200$ nodes we ran experiments on a machine with 16GB RAM and 8 processors Intel(R) Core(TM) i5-8265U CPU at 1.60GHz. For $500$ or more nodes we used a machine with 256GB RAM and 64 processors AMD EPYC 7301 16-Core Processor at 2.20GHz.\\[.5em]
The metrics used are precision, recall, Structural Hamming Distance (SHD) -- which is computed as the sum of false positive, false negative and wrongly directed edges -- and Structural Intervention Distance (SID) (\cite{sid_2013}) -- accounting for the number of miscalculated interventional distributions that would result from the inferred graph.\\[.5em]
We focus our experiments on synthetic data to show the scalability properties of DAS with increasing number of nodes. In order to sample data from the nonlinear additive Gaussian noise model of Equation \eqref{eq:model}  we mimic the experimental setting of \cite{rolland_2022}. The causal graphs are generated using the Erd\"{o}s-Renyi model (\cite{Erdos:1960}). We run experiments fixing the number of nodes $d$ as well as the sparsity of the graph by setting the expected amount of edges to be equal to $d$ (ER1) or $4d$ (ER4). For $d > 200$ we drop the SID metric as it is too slow to compute. Whenever results for some method are not appearing in the table this means we could not perform these runs in a reasonable time. We repeat the experiments for 10 times and report empirical mean and standard deviation over the metrics. The number of samples is maintained fixed at $n = 1000$ and we set $K=20$ (same value found SCORE and CAM experimental settings).
\begin{figure}[]
    \centering
    \includegraphics[width=.49\textwidth]{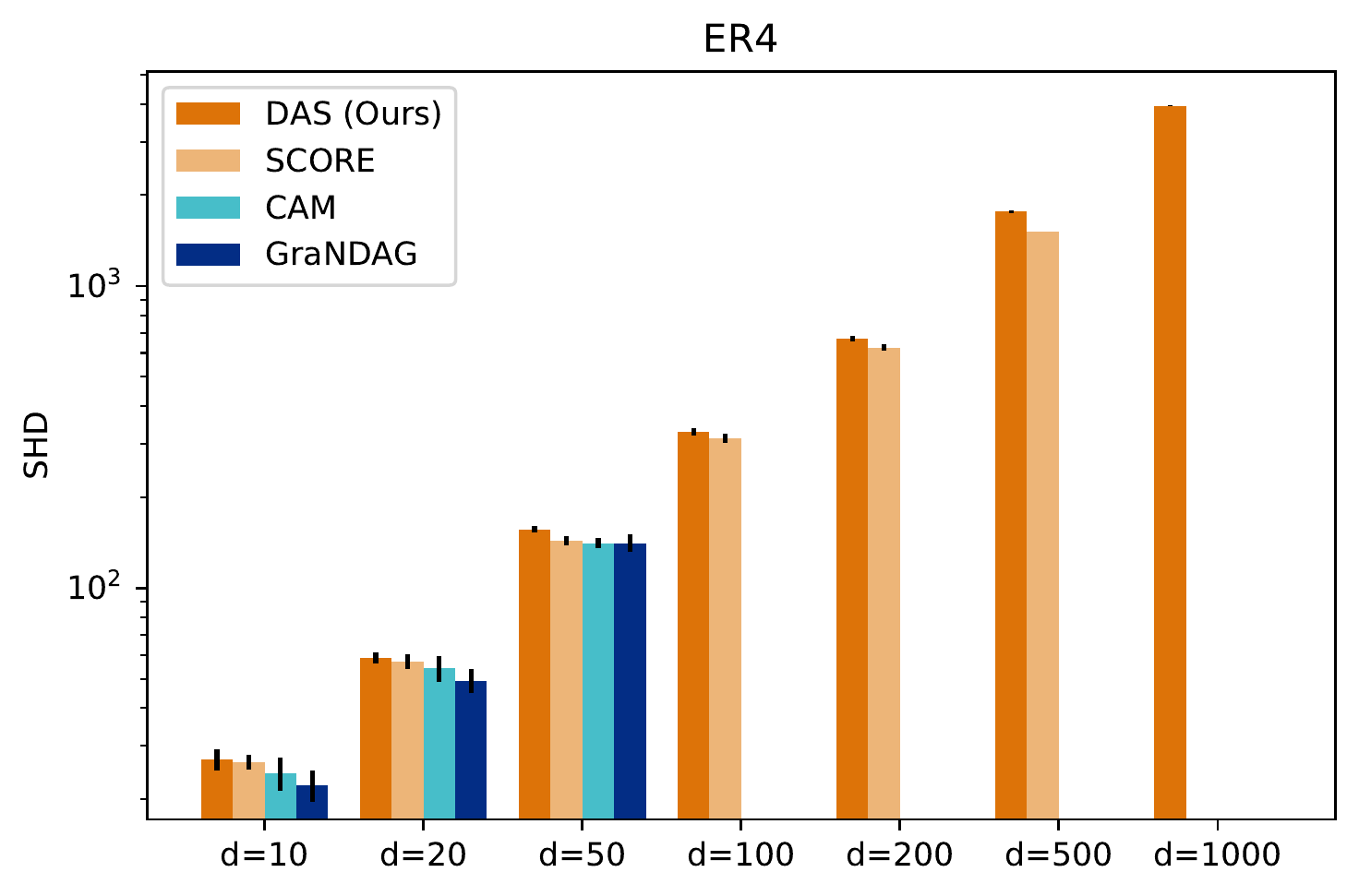}
    \includegraphics[width=.49\textwidth]{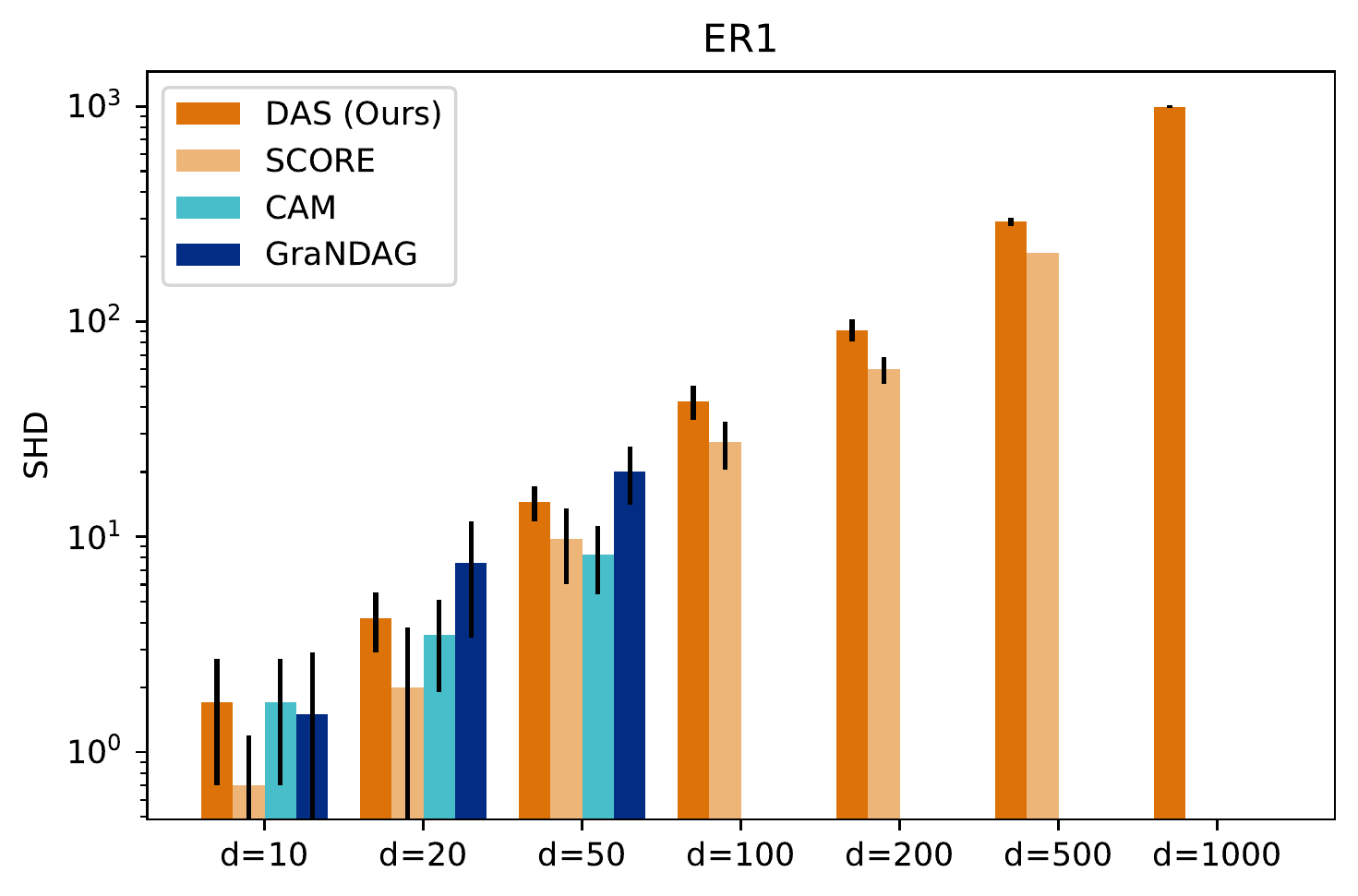}
    \caption{SHD versus $d$ number of nodes for different methods on dense (left) and sparse (right) graphs. For higher values of $d$ some methods are missing as they were too much time expensive to run. Number of samples is $n = 1000$.}
    \label{fig:shd2}
\end{figure}
\\[.5em]
\looseness-1From Table \ref{tab:er4} we can see that on denser graphs (ER4) our method maintains similar performance with respect to the other three for nodes up to 50, while being considerably faster in particular with respect to GraNDAG and CAM. As $d$ increases, the gap with SCORE reduces up to the point that for $200$ nodes we observe better SID for our algorithm. At $d \geq 500$ it becomes arguably impossible to run SCORE on a personal computer in a finite amount of time, whereas DAS is the only reasonable option.\\
Similarly, the performances across the different methods are comparable when running inference on sparser graphs (ER1), as reported in Table \ref{tab:er1}. These results are directly observable in Figure \ref{fig:shd2}: each algorithm shows a similar degrade in performance with the number of nodes increasing, and bars set to close SHD values. Nevertheless, in Figure \ref{fig:time} it clearly appears that DAS achieves these metrics in a significantly smaller amount of time, supporting the claim of better efficiency in terms of velocity and scalability of our approach.\\[.4em]
In Appendix \ref{app:das_alpha_stability} we provide additional empirical results, focused on the stability of DAS with respect to changes in the threshold for hypothesis testing of mean larger than zero. Moreover, we significantly extend our experiments testing DAS on Scale-free synthetic graphs (\cite{Barabasi99emergenceScaling}, Appendix \ref{app:sf_experiments}), and on Sachs real data (\cite{sachs_2005}) and semi-synthetic data sampled from SynTReN generator (\cite{syntren_2006})), Appendix \ref{app:sachs_syntren}).

\begin{table}[!t]
\footnotesize
\caption{Experiments on ER4 data. For CAM and GraNDAG we report results found in \cite{rolland_2022}. For higher values of $d$ some methods are missing as they were too much time expensive to run.}
\label{tab:er4}
\centering
\begin{tabular}{clccccc}
\toprule
 & Method & SHD & SID & Prec. & Rec. & Time [s]\\
\midrule
\multirow{4}{*}{d=10}   & DAS (Ours) & $27.0 \pm 2.2$ & $43.6 \pm 5.8$ & $1.00 \pm 0.00$ & $0.33 \pm 0.02$ & $\mathbf{7.7 \pm 0.1}$\\
& SCORE & $26.5 \pm 1.5$ & $42.3 \pm 2.9$ & $0.99 \pm 0.00$ & $0.33 \pm 0.02$ & $7.9 \pm 0.1$\\
& CAM & $24.4 \pm 3.1$ & $45.2 \pm 10.2$ & $-$ & $-$ &  $30.1 \pm 3.7$\\
& GraNDAG & $\mathbf{22.2 \pm 2.6}$ & $\mathbf{42.0 \pm 6.2}$ & $-$ & $-$ &  $185 \pm 26$\\
\midrule
\multirow{4}{*}{d=20}  & DAS (Ours) & $56.4 \pm 2.5$ & $213 \pm 28$  & $0.99 \pm 0.00$ & $0.27 \pm 0.04$ & $\mathbf{16.1 \pm 0.3}$\\
& SCORE & $57.17 \pm 3.1$ & $229 \pm 23$ & $0.99 \pm 0.01$ & $0.30 \pm 0.04$ & $40.7 \pm 1.8$\\
& CAM & $54.2 \pm 5.4$ & $202 \pm 29$ & $-$ & $-$ & $313 \pm 80$\\
& GraNDAG & $\mathbf{49.3 \pm 4.5}$ & $\mathbf{211 \pm 37}$ & $-$ & $-$ & $357 \pm 47$\\
\midrule
\multirow{4}{*}{d=50}  & DAS (Ours) & $156\pm 4$ & $1460 \pm 67$  & $0.96 \pm 0.02$ & $0.24 \pm 0.03$ & $\mathbf{48.3 \pm 1.1}$\\
& SCORE & $144 \pm 6$ & $1346 \pm 57$ & $0.97 \pm 0.01$ & $0.30 \pm 0.03$ & $245 \pm 5$\\
& CAM & $\mathbf{141 \pm 6}$ & $\mathbf{1337 \pm 94}$ & $-$ & $-$ & $1143 \pm 79$\\
& GraNDAG & $\mathbf{141 \pm 10}$ & $1432 \pm 110$ & $-$  & $-$ & $1410 \pm 73$\\
\midrule
\multirow{2}{*}{d=100}  & DAS (Ours) & $335 \pm 5$ & $6695 \pm 224$ & $0.91 \pm 0.03$ & $0.21 \pm 0.04$ & $\mathbf{114 \pm 2}$\\
& SCORE & $\mathbf{313 \pm 11}$ & $\mathbf{5965 \pm 273}$ & $0.91 \pm 0.03$ & $0.27 \pm 0.06$ &  $779 \pm 13$\\
\midrule
\multirow{2}{*}{d=200}  & DAS (Ours) & $691 \pm 13$ & $\mathbf{25147 \pm 835}$  & $0.89 \pm 0.06$ & $0.21 \pm 0.05$ & $\mathbf{279 \pm 4}$\\
& SCORE & $\mathbf{626 \pm 14}$ & $25707 \pm 891$ & $0.88 \pm 0.04$ & $0.30 \pm 0.05$ & $4142 \pm 35$\\
\midrule
\multirow{2}{*}{d=500}  & DAS (Ours) & $1761 \pm 15$ & $-$ & $0.80 \pm 0.04$ & $0.19 \pm 0.03$ & $\mathbf{1308 \pm 7}$\\
& SCORE\footnote{} & $\mathbf{1642}$ & $-$ & $0.82$ & $0.27$ & $25307$\\
\midrule
\multirow{1}{*}{d=1000}  & DAS (Ours) & $\mathbf{3951 \pm 9}$ & $-$ & $0.76 \pm 0.05$ & $0.08 \pm 0.00$ & $\mathbf{5539 \pm 81}$\\
\bottomrule
\end{tabular}
\begin{flushleft}
\small{$^2$ For $d=500$ and SCORE method no standard deviation appears because experiments could not be repeated in a reasonable time. The values in the table refer to a single run.}
\end{flushleft}
\end{table}

\begin{table}[h!]
\footnotesize
\caption{Experiments on ER1 data. For CAM and GraNDAG we report results found in \cite{rolland_2022}. For higher values of $d$ some methods are missing as they were too much time expensive to run.}
\label{tab:er1}
\centering
\begin{tabular}{clccccc}
\toprule
 & Method & SHD & SID & Prec. & Rec. & Time [s]\\
\midrule
\multirow{4}{*}{d=10}   & DAS (Ours) & $1.0 \pm 0.8$ & $\mathbf{3.3 \pm 3.7}$ & $0.97 \pm 0.01$ & $0.84 \pm 0.05$ & $\mathbf{7.8 \pm 0.1}$\\
& SCORE & $\mathbf{0.7 \pm 0.5}$ & $4.5 \pm 4.3$ & $0.98 \pm 0.01$ & $0.98 \pm 0.01$ & $8.0 \pm 0.2$\\
& CAM & $1.7 \pm 1.0$ & $6.4 \pm 4.2$ & $-$ & $-$ & $30.1 \pm 3.7$\\
& GraNDAG & $1.5 \pm 1.4$ & $6.5 \pm 7.2$ & $-$ & $-$ & $185 \pm 26$\\
\midrule
\multirow{4}{*}{d=20}  & DAS (Ours) & $3.2 \pm 1.4$ & $18.1 \pm 10.1$  & $0.98 \pm 0.02$ & $0.85 \pm 0.03$ & $\mathbf{16.7 \pm 0.4}$\\
& SCORE & $\mathbf{2.0 \pm 1.8}$ & $\mathbf{8.3 \pm 9.9}$ & $0.99 \pm 0.01$ & $0.91 \pm 0.03$ & $36.4 \pm 1.8$\\
& CAM & $3.5\pm 1.6$ & $14.3 \pm 9.8$ & $-$ & $-$ & $313 \pm 80$\\
& GraNDAG & $7.6 \pm 4.2$ & $31.6 \pm 22.7$ & $-$ & $-$ & $357 \pm 47$\\
\midrule
\multirow{4}{*}{d=50}  & DAS (Ours) & $17.3 \pm 3.4$ & $99.6 \pm 42.1$  & $0.96 \pm 0.04$ & $0.77 \pm 0.04$  & $\mathbf{49.2 \pm 2.1}$\\
& SCORE & $9.8 \pm 3.8$ & $69.6 \pm 41.3$ & $0.98 \pm 0.01$ & $0.87 \pm 0.03$ & $251 \pm 7$\\
& CAM & $\mathbf{8.3 \pm 2.9}$ & $\mathbf{53.7 \pm 31.9}$ & $-$ & $-$ & $1143 \pm 79$\\
& GraNDAG & $20.2 \pm 6.1$ & $135 \pm 456$ & $-$ & $-$ & $1410 \pm 73$\\
\midrule
\multirow{2}{*}{d=100}  & DAS (Ours) & $46.6 \pm 9.1$ & $327 \pm 47$ & $0.92 \pm 0.06$ & $0.68 \pm 0.04$ & $\mathbf{111 \pm 6}$\\
& SCORE & $\mathbf{27.5 \pm 6.9}$ & $\mathbf{288 \pm 115}$ & $0.97 \pm 0.02$ & $0.83 \pm 0.05$ & $776 \pm 12$\\
\midrule
\multirow{2}{*}{d=200}  & DAS (Ours) & $110 \pm 6.6$ & $899 \pm 191$  & $0.88 \pm 0.07$ & $0.68 \pm 0.06$ & $\mathbf{282 \pm 4}$\\
& SCORE & $\mathbf{59.9 \pm 8.5}$ & $\mathbf{495 \pm 161}$ & $0.95 \pm 0.03$ & $0.85 \pm 0.07$ & $4237 \pm 22$\\
\midrule
\multirow{2}{*}{d=500}  & DAS (Ours) & $291 \pm 13$ & $-$ & $0.78 \pm 0.07$ & $0.65 \pm 0.05$ & $\mathbf{1329 \pm 7}$\\
& SCORE\footnote{For $d=500$ and SCORE method no standard deviation appears because experiments could not be repeated in a reasonable time. The values in the table refer to a single run.} & $\mathbf{209}$ & $-$ & $0.8$ & $0.85$  & $25115$\\
\midrule
\multirow{1}{*}{d=1000}  & DAS (Ours) & $\mathbf{994 \pm 15}$ & $-$ & $0.59 \pm 0.02$ & $0.09 \pm 0.00$ & $\mathbf{5544 \pm 73}$\\
\bottomrule
\end{tabular}
\begin{flushleft}
\small{$^3$ For $d=500$ and SCORE method no standard deviation appears because experiments could not be repeated in a reasonable time. The values in the table refer to a single run.}
\end{flushleft}
\end{table}

\section{Conclusion}
Under the assumption of nonlinear additive Gaussian noise model of the data we showed how to theoretically recover the exact causal graph from the Jacobian of the score function. Our finding extends the work of \cite{rolland_2022}, using the score to learn the topological ordering of the variables from the data. In addition to this, we showed that all edges can be discovered and oriented from the score. Based on our analysis, we designed an algorithm that is more efficient by a factor of $\mathcal{O}(d)$ compared to~\cite{rolland_2022}, yielding significant speed up in practice while retaining comparable accuracy.\\[.5em]
While we already obtained a significant speed up, further improvement could be achieved by amortizing the score computation over subgraphs or even multiple graphs~(\cite{lowe2020amortized}) or using more scalable estimators. However, the main blocker for future work (and biggest limitation of our work) is the lack of a public, highly curated, and large benchmark for causal discovery. In fact, our scalability experiments were limited to inference on synthetic data. Without real data it is hard to evaluate algorithms' performance outside of a controlled synthetic setting. As huge and fully annotated causal graphs may never become available, we identify three important directions for future work. First, extending our approach to input variables outside of the model assumptions, in particular considering data coming from interventions and with latent confounders. Second, evaluation protocols with partially annotated causal graphs and externally collected interventional distributions. Third, merging many smaller and annotated data sets that share similar but non-overlapping variables~(\cite{mejia2022obtaining}).

\acks{We want to thank Volkan Cevher for the valuable discussions. This work has been supported by  AFOSR,  grant n. FA8655-20-1-7035. FM is supported by \textit{Programma Operativo Nazionale ricerca e innovazione 2014-2020}.}
\bibliography{biblio}

\newpage
\appendix

\section{Proof of Lemma \ref{lem:lemma1}}\label{app:proof}
In this section we provide a proof of the statement of Lemma \ref{lem:lemma1} for completeness.

\begin{proof}
\hspace{.4em} For a leaf $l$ the score of Equation \eqref{eq:score_entry} becomes $s_l(\mathbf{X}) = -\frac{X_l - f_l(\operatorname{pa}_l(\mathbf{X}))}{\sigma_l^2}$. We compute the partial derivative 
\begin{equation}
    \frac{\partial s_l(\mathbf{X})}{\partial X_j} = \frac{1}{\sigma_l^2} \frac{\partial f_l}{\partial X_j}(\operatorname{pa}_l(\mathbf{X}))
    \label{eq:proof}
\end{equation}
and observe that:

\begin{enumerate}[label=(\roman*)]
    \item $\mathbf{E}\left[ \abs*{\frac{\partial s_l(\mathbf{X})}{\partial X_j}}\right] \neq 0 \Rightarrow X_j \in \operatorname{pa}_l(\mathbf{X})$. By contradiction, consider $X_j \not\in \operatorname{pa}_l(\mathbf{X})$: being $f_l(\operatorname{pa}_l(\mathbf{X}))$ constant in $X_j$, then $\frac{\partial f_l(\operatorname{pa}_l(\mathbf{X}))}{\partial X_j} = 0$ for every $\mathbf{X} \in \mathbb{R}^d$ by definition of derivative. Then, $\mathbf{E}\left[ \abs*{\frac{\partial s_l(\mathbf{X})}{\partial X_j}}\right] = 0$, which contradicts the hypothesis.
    \item $X_j \in \operatorname{pa}_l(\mathbf{X}) \Rightarrow \mathbf{E}\left[ \abs*{\frac{\partial s_l(\mathbf{X})}{\partial X_j}}\right] \neq 0$: we observe from Equation \eqref{eq:method} that $\frac{\partial f_l}{\partial X_j}(\operatorname{pa}_l(\mathbf{X})) \neq 0$ \textit{almost surely}, such that $\abs*{\frac{\partial f_l}{\partial X_j}(\operatorname{pa}_l(\mathbf{X}))} > 0$ \textit{almost surely}. Being the probability of vanishing $\abs*{\frac{\partial f_l}{\partial X_j}(\operatorname{pa}_l(\mathbf{X}))}$ equals to zero, then the expectation $\mathbf{E}\left[ \abs*{\frac{\partial s_l(\mathbf{X})}{\partial X_j}}\right]$ is equivalent to the integral $\mathlarger{\int}_{\mathcal{X}^+} \abs*{\frac{\partial f_l}{\partial X_j}(\operatorname{pa}_l(\mathbf{X}))} dP(\mathbf{X})$, with $\mathcal{X}^+ \subseteq \mathbb{R}^d$ the subset of values where $\abs*{\frac{\partial f_l}{\partial X_j}(\operatorname{pa}_l(\mathbf{X}))}$ is strictly positive. Since the integral of a strictly positive function is strictly positive itself, then $\mathbf{E}\left[ \abs*{\frac{\partial s_l(\mathbf{X})}{\partial X_j}}\right] > 0$.
\end{enumerate}\end{proof}


\section{DAS stability with respect to hypothesis testing threshold}\label{app:das_alpha_stability}
In our experiments of Section \ref{sec:exp}, we fix the $\alpha$ threshold for hypothesis testing of non-zero mean to $0.01$. Standard $\alpha$ values are $0.1$ or lower. In absence of specific information on the data, most of causal discovery methods based on conditional independence testing employ default threshold of $0.05$ (for instance, see FCI and PC implementations on \texttt{DoDiscover} and \texttt{causal-learn} well known libraries): for comparison, in Table \ref{tab:alpha_experiments} we provide experimental results of DAS with $\alpha$ threshold set to $0.05$. We run $10$ experiments with different seeds and report empirical mean and standard deviation. Experiments show that $\alpha=0.05$ doesn't significantly affect the performance of DAS (distance to the average SHD with $\alpha=0.01$ is always within error bars). 
\begin{table}[]
    \caption{DAS experiment on ER1 and ER4 data with different $\alpha$ cutoff thresholds for hypothesis testing. }
    \label{tab:alpha_experiments}
    \centering
    \begin{tabular}{lccc}
        \toprule
         Nodes & Threshold & SHD (ER1) & SHD (ER4) \\
         \midrule
         \multirow{2}{*}{$d=10$} & $\alpha=0.01$ & $1.0 \pm 0.8$ & $27.0 \pm 2.2$ \\
         & $\alpha=0.05$ & $0.9 \pm 0.9$ & $26.7 \pm 2.0$ \\
         \midrule
         \multirow{2}{*}{$d=20$} & $\alpha=0.01$ & $3.2 \pm 1.4$ & $56.4 \pm 2.5$ \\
         & $\alpha=0.05$ & $3.1 \pm 1.6$ & $58.2 \pm 2.5$ \\
         \midrule
         \multirow{2}{*}{$d=50$} & $\alpha=0.01$ & $17.3 \pm 3.4$ & $156 \pm 4$ \\
         & $\alpha=0.05$ & $17.1 \pm 3.7$ & $155 \pm 5$ \\
         \midrule
         \multirow{2}{*}{$d=100$} & $\alpha=0.01$ & $46.6 \pm 9.1$ & $355 \pm 5$ \\
         & $\alpha=0.05$ & $47.6 \pm 5.3$ & $339 \pm 6$ \\
         \midrule
         \multirow{2}{*}{$d=200$} & $\alpha=0.01$ & $110 \pm 6$ & $691 \pm 13$ \\
         & $\alpha=0.05$ & $114 \pm 8$ & $698 \pm 9$ \\
         \bottomrule
    \end{tabular}
\end{table}


\section{SF Experiments}\label{app:sf_experiments}
In this section we present experimental results on Scale Free graphs, both on sparser (Table \ref{tab:sf1}) and denser graphs (Table \ref{tab:sf4}).

\begin{table}[h!]
\footnotesize
\caption{Experiments on SF1 data. For SCORE, CAM and GraNDAG we report results found in \cite{rolland_2022}.}
\label{tab:sf1}
\centering
\begin{tabular}{clccccc}
\toprule
 & Method & SHD & SID & Prec. & Rec. & Time [s]\\
\midrule
\multirow{4}{*}{d=10}   & DAS (Ours) & $0.4 \pm 0.7$ & $2.2 \pm 3.4$ & $0.99 \pm 0.04$ & $0.84 \pm 0.15$ & $7.6 \pm 0.1$\\
& SCORE & $0.3 \pm 0.6$ & $ 2.7 \pm 5.8$ & $-$ & $-$ & $-$\\
& CAM & $0.4 \pm 0.5$ & $2.8 \pm 3.6$ & $-$ & $-$ & $-$\\
& GraNDAG & $1.4 \pm 1.0$ & $12.5 \pm 9.7 $ & $-$ & $-$ & $-$\\
\midrule
\multirow{4}{*}{d=20}  & DAS (Ours) & $1.9 \pm 1.6$ & $19.1 \pm 7.4$ & $0.99 \pm 0.02$ & $0.84 \pm 0.11$ & $16.6 \pm 0.4$\\
& SCORE & $0.9 \pm 0.9$ & $13.8 \pm 12.6$ & $-$ & $-$ & $-$\\
& CAM & $0.9 \pm 0.9$ & $12.9 \pm 14.0$ & $-$ & $-$ & $-$\\
& GraNDAG & $3.2 \pm 1.9$ & $25.5 \pm 15.6$ & $-$ & $-$ & $-$\\
\midrule
\multirow{4}{*}{d=50\footnote{}}  & DAS (Ours) & $10.9 \pm 4.9$ & $225.3 \pm 72.1$  & $0.96 \pm 0.03$ & $0.74 \pm 0.08$ & $53.1 \pm 0.9$\\
& SCORE & $4.6 \pm 2.4$ & $132.6 \pm 75.8$ & $-$ & $-$ & $-$\\
& CAM & $3.6 \pm 1.9$ & $ 115.4 \pm 72.6$ & $-$ & $-$ & $-$\\
& GraNDAG & $9.2 \pm 3.3$ & $281.8 \pm 129.8$ & $-$ & $-$ & $-$\\
\midrule
\multirow{1}{*}{d=100}  & DAS (Ours) & $39.4 \pm 8.0$ & $214 \pm 42$ & $0.94 \pm 0.02$ & $0.51 \pm 0.07$ & $102 \pm 5$\\
\midrule
\multirow{1}{*}{d=200}  & DAS (Ours) & $112 \pm 11$ & $626 \pm 83$  & $0.97 \pm 0.02$ & $0.33 \pm 0.03$ & $279 \pm 6$\\
\midrule
\multirow{1}{*}{d=500}  & DAS (Ours) & $271 \pm 15$ & $-$ & $0.93 \pm 0.09$ & $0.68 \pm 0.01$ & $1315 \pm 9$\\
\midrule
\multirow{1}{*}{d=1000}  & DAS (Ours) & $910 \pm 12$ & $-$ & $0.59 \pm 0.02$ & $0.09 \pm 0.00$ & $5442 \pm 61$\\
\bottomrule
\end{tabular}
\begin{flushleft}
\small{$^4$ For $d > 50$  experiments are executed only for DAS.}
\end{flushleft}
\label{tab:sf1}
\end{table}

\begin{table}[h!]
\footnotesize
\caption{Experiments on SF4 data. For SCORE, CAM and GraNDAG we report results found in \cite{rolland_2022}.}
\label{tab:sf4}
\centering
\begin{tabular}{clccccc}
\toprule
 & Method & SHD & SID & Prec. & Rec. & Time [s]\\
\midrule
\multirow{4}{*}{d=10}   & DAS (Ours) & $10.1 \pm 2.39$ & $35.7 \pm 9.1$ & $0.99 \pm 0.01$ & $0.75 \pm 0.06$ & $7.8 \pm 0.1$\\
& SCORE & $ 4.6 \pm 1.7$ & $21.5 \pm 9.6$ & $-$ & $-$ & $-$\\
& CAM & $9.6 \pm 2.0$ & $40.4 \pm 11.4$ & $-$ & $-$ & $-$\\
& GraNDAG & $4.7 \pm 1.8$ & $23.0 \pm 7.3$ & $-$ & $-$ & $-$\\
\midrule
\multirow{4}{*}{d=20}  & DAS (Ours) & $30.4 \pm 5.78$ & $248.1 \pm 20.0$ & $0.98 \pm 0.02$ & $0.59 \pm 0.06$ & $17.1 \pm 0.7$\\
& SCORE & $17.5 \pm 3.5$ & $ 179.2 \pm 23.8$ & $-$ & $-$ & $-$\\
& CAM & $26.4 \pm 3.9$ & $ 253.7 \pm 28.8$ & $-$ & $-$ & $-$\\
& GraNDAG & $14.7 \pm 4.0$ & $168.0 \pm 39.2$ & $-$ & $-$ & $-$\\
\midrule
\multirow{4}{*}{d=50\footnote{For $d > 50$  experiments are executed only for DAS.}}  & DAS (Ours) & $115.5 \pm 10.8$ & $703.1 \pm 87.5$  & $0.97 \pm 0.01$ & $0.45 \pm 0.06$ & $51.3 \pm 1.2$\\
& SCORE & $68.3 \pm 3.6$ & $1724 \pm 109$ & $-$ & $-$ & $-$\\
& CAM & $85.3 \pm 4.2$ & $1935 \pm 99$ & $-$ & $-$ & $-$\\
& GraNDAG & $63.8 \pm 9.7$ & $1677 \pm 118$ & $-$ & $-$ & $-$\\
\midrule
\multirow{1}{*}{d=100}  & DAS (Ours) & $295.3 \pm 10.8$ & $3212 \pm 145$ & $0.97 \pm 0.01$ & $0.25 \pm 0.02$ & $108 \pm 2$\\
\midrule
\multirow{1}{*}{d=200}  & DAS (Ours) & $674.2 \pm 16.6$ & $21314 \pm 891$  & $0.95 \pm 0.02$ & $0.11 \pm 0.01$ & $283 \pm 3$\\
\midrule
\multirow{1}{*}{d=500}  & DAS (Ours) & $1890 \pm 7.1$ & $-$ & $0.97 \pm 0.03$ & $0.02 \pm 0.01$ & $1212 \pm 31$\\
\midrule
\multirow{1}{*}{d=1000}  & DAS (Ours) & $3715 \pm 21.3$ & $-$ & $0.92 \pm 0.04$ & $0.03 \pm 0.01$ & $5016 \pm 53$\\
\bottomrule
\end{tabular}
\begin{flushleft}
\small{$^5$ For $d > 50$  experiments are executed only for DAS.}
\end{flushleft}
\label{tab:sf4}
\end{table}


\section{Real and semi-synthetic data}\label{app:sachs_syntren}
In addition to experiments on synthetic graphs, we test the empirical performance of DAS on Sachs real data (\cite{sachs_2005}, a common benchmark for causal discovery on biological data with 17 edges and 853 observations)
and on semi-synthetic data sampled from SynTReN generator of gene expression records (\cite{syntren_2006}). 
In Table \ref{tab:sachs} we see that DAS matches SCORE performance on Sachs, whereas it is comparable to SCORE, CAM and GraNDAG on SynTReN experiments 
(Table \ref{tab:syntren}).
\begin{table}[h!]
    \caption{Experiments on Sachs dataset.}
    \label{tab:sachs}
    \centering
    \begin{tabular}{lcc}
    \toprule
     Method & SHD & SID\\
     \midrule
     DAS (Ours) & $12$ & $45$ \\
     SCORE  & $12$ & $45$ \\
     CAM  & $12$ & $55$ \\
     GraNDAG & $13$ & $47$ \\
    \end{tabular}
\end{table}

\begin{table}[h!]
    \caption{Experiments on SynTReN data (20 nodes, 500 samples). Empirical mean and error are computed over $10$ runs on different seeds.}
    \label{tab:syntren}
    \centering
    \begin{tabular}{lcc}
    \toprule
     Method & SHD & SID\\
     \midrule
     DAS (Ours) & $38.0 \pm 4.1$ & $188.5 \pm 71.2$ \\
     SCORE  & $36.8 \pm 4.7$ & $193.4 \pm 60.2$ \\
     CAM  & $40.5 \pm 6.8$ & $152.3 \pm 48.0$ \\
     GraNDAG & $34.0 \pm 8.5$ & $161.7 \pm 53.4$ \\
    \end{tabular}
\end{table}
\end{document}